\documentclass[letterpaper, 10 pt, conference]{ieeeconf}  

\IEEEoverridecommandlockouts                              

\usepackage{amsmath,amssymb,amsfonts}
\usepackage{siunitx}
\usepackage{algorithmic}
\usepackage{graphicx}
\usepackage{textcomp}
\usepackage{pgfplots}
\pgfplotsset{compat=newest}
\usepackage{comment}
\usepackage{tikz}
\usepackage{subcaption}
\usepackage{authblk}
\usepackage{booktabs}
\usepackage{xcolor}
\usepackage[normalem]{ulem}

\definecolor{new_blue}{HTML}{1f77b4}
\definecolor{new_orange}{HTML}{ff7f0e}
\definecolor{new_green}{HTML}{2ca02c}
\definecolor{new_red}{HTML}{d62728}
\definecolor{new_purple}{HTML}{9467bd}

\title{\LARGE \bf
Towards Generalizing Sensorimotor Control Across Weather Conditions
}

\author{Qadeer Khan$^{1,2*}$ \ \ \
Patrick Wenzel$^{1,2*}$ \ \ \
Daniel Cremers$^{1,2}$ \ \ \
Laura Leal-Taix\'{e}$^{1}$
\thanks{$^{*}$These authors contributed equally.}
\thanks{$^{1}$Technical University of Munich}
\thanks{$^{2}$Artisense}
}

\begin{document}
\maketitle
\thispagestyle{empty}
\pagestyle{empty}


\begin{abstract}
The ability of deep learning models to generalize well across different scenarios depends primarily on the quality and quantity of annotated data. Labeling large amounts of data for all possible scenarios that a model may encounter would not be feasible; if even possible. We propose a framework to deal with limited labeled training data and demonstrate it on the application of vision-based vehicle control. We show how limited steering angle data available for only one condition can be transferred to multiple different weather scenarios. This is done by leveraging unlabeled images in a teacher-student learning paradigm complemented with an image-to-image translation network. The translation network transfers the images to a new domain, whereas the teacher provides soft supervised targets to train the student on this domain. Furthermore, we demonstrate how utilization of auxiliary networks can reduce the size of a model at inference time, without affecting the accuracy. The experiments show that our approach generalizes well across multiple different weather conditions using only ground truth labels from one domain.
\end{abstract}


\section{Introduction}\label{sec:introduction}

The ubiquity of a tremendous amount of processing power in contemporary computing units has proliferated the usage of deep learning-based approaches in control applications. In particular, supervised deep learning methods have made great strides in sensorimotor control, whether it be for autonomous driving~\cite{BojarskiArXiv2016}, robot perception~\cite{KaufmannCoRL2018}, or manipulation tasks~\cite{LevineJMLR2016, NairICRA2017, ZhangICRA2018}. However, the performance of such models is heavily dependent on the availability of ground truth labels. To have the best generalization capability, one should annotate data for all possible scenarios. Nonetheless, obtaining labels of high quality is a tedious, time consuming, and error-prone process. 

\begin{figure}
  \centering
  \includegraphics[width=0.75\linewidth]{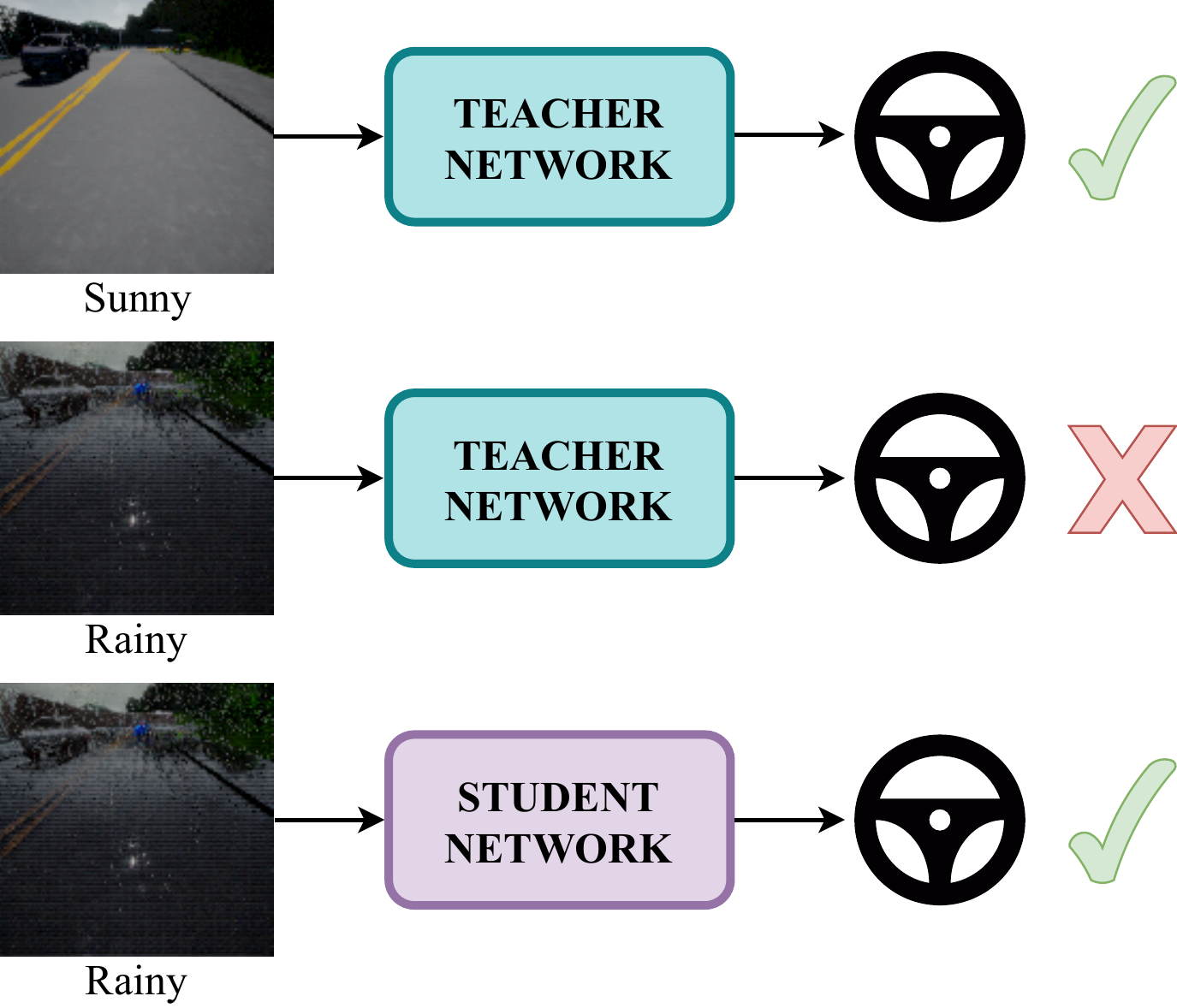}
  \caption{Teacher-student training for generalizing sensorimotor control across weather conditions. \textbf{Top:} The teacher network, trained on ground truth data collected on sunny weather is capable of predicting the correct steering angle when tested on this condition. \textbf{Middle:} However, the teacher fails to predict the correct steering when tested on an input image from a different domain (rainy weather). \textbf{Bottom:} With our proposed framework, the student network trained with supervised information from the teacher network is capable of predicting the correct steering for the rainy weather. This is done without any additional ground truth labels or semantic information.}
  \label{fig:intro_fig}
\end{figure}

We propose to instead utilize the information available for one domain and transfer it to a different one without human supervision as shown in Figure~\ref{fig:intro_fig}. This is particularly helpful for many robotic applications wherein a robotic system trained in one environment should generalize across different environments without human intervention. For example in simultaneous localization and mapping (SLAM), it is very important that the algorithm is robust to different lighting conditions~\cite{NewmanCVPR2017}. 
In the context of autonomous driving, transferring knowledge from simulation to the real world or between different weather conditions is of high relevance. Recently,~\cite{MullerCoRL2018, YouBMVC2017, WenzelCoRL2018} have attempted to tackle these problems by dividing the task of vehicle control into different modules, where each module specialized in extracting features from a particular domain. In these works, semantic labels are used as an intermediate representation for transferring knowledge between different domains. However, obtaining these semantic labels requires human effort which is time-consuming, expensive, and error-prone \cite{WenzelCoRL2018}. In this work, we instead propose to use a teacher-student learning-based approach to generalize sensorimotor control across weather conditions without the need for extra annotations, \emph{e.g.}, semantic segmentation labels. 

To this end, we make the following contributions:

\begin{itemize}
\item We demonstrate how knowledge of ground truth data for steering angles can be transferred from one weather scenario to multiple different weather conditions. This is achieved without the additional requirement of having semantic labels. We make use of an image-to-image translation network to transfer the images between different domains while preserving information necessary for taking a driving decision.
\item We show how the proposed method can also utilize images without ground truth steering commands to train the models using a teacher-student framework. The teacher provides relevant supervised information regarding the unlabeled images to train the features of the student. Hence, we can eliminate the need for an expert driver for data collection across diverse conditions.
\item If the sample data with ground truth labels is limited, then the teacher and student models may tend to overfit. To overcome this, we propose using weighted auxiliary networks connected to the intermediate layers of these models. During inference, the model size can be reduced by eliminating auxiliary layers with low weights without reducing accuracy. 
\end{itemize}

In the following sections, we first review related work. We then present the details of our method, followed by an analysis of our model's performance. Finally, we discuss various parts of our model.

\section{Related Work}\label{sec:related_work}

Vision-based autonomous driving approaches have been studied extensively in an academic and industrial setting~\cite{JanaiArXiv2017}. A plenty of real world~\cite{GeigerCVPR2012,CordtsCVPR2016,XuCVPR2017} as well as synthetic~\cite{RosCVPR2016,GaidonCVPR2016,RichterICCV2017,RichterECCV2016} datasets for autonomous driving research have become available. In recent years, neural network approaches have significantly advanced the state-of-the-art in computer vision tasks. Especially, end-to-end learning for sensorimotor control has recently gained a lot of interest in the vision and robotics community. In this context, different approaches to autonomous driving are studied: modular pipelines~\cite{ThrunJFR2006}, imitation learning~\cite{PomerleauNIPS1989}, conditional imitation learning~\cite{Codevilla2018ICRA}, and direct perception~\cite{ChenICCV2015}.

\noindent{\textbf{Embodied agent evaluation. }}Most available datasets~\cite{GeigerCVPR2012, CordtsCVPR2016} cannot be used for evaluating online driving performance due to their static nature. The evaluation of driving models on realistic data is challenging and often not feasible. Therefore, a lot of interest has emerged in building photo-realistic simulators~\cite{MullerIJCV2018, ShahFSR2017, DosovitskiyCoRL2017} to analyze those models. However, despite having access to simulation engines, there is currently no universally accepted benchmark to evaluate vision-based control agents. Therefore, our experimental setup is a step towards a field where it is still not quite established how to evaluate and measure the performance of the models~\cite{AndersonArXiv2018, CodevillaECCV2018}. 

\noindent{\textbf{Unpaired image-to-image translation networks. }}Unsupervised image-to-image translation techniques are rapidly making progress in generating high-fidelity images across various domains~\cite{ZhuICCV2017, LiuNIPS2017, HuangECCV2018, MinjunECCV2018}. Our framework is agnostic to any particular method. Hence, continual improvements in these networks can be easily integrated into our framework by replacing a previous network.

\noindent{\textbf{Transfer learning via semantic modularity. }}Several works used semantic labels of the scene as an 
intermediate representation for transferring knowledge between domains. In the context of autonomous driving, the authors of~\cite{MullerCoRL2018} proposed to map the driving policy utilizing semantic segmentation to a local trajectory plan to be able to transfer between simulation and real-world data. Furthermore, for making a reinforcement model trained in a virtual environment workable in the real world, the authors of~\cite{YouBMVC2017} utilize the intermediate semantic representation as well to translate virtual to real images. However, there is still little work on generalizing driving models across weathers. The work by~\cite{WenzelCoRL2018} showed how to transfer knowledge between different weather conditions using a semantic map of the scene. In contrast, in this paper, we demonstrate the possibility of transferring the knowledge between weathers even without semantic labels. 

\noindent{\textbf{Knowledge distillation. }}Originally, knowledge distillation~\cite{HintonArXiv2015} was used for network compression (student network is smaller than the teacher while maintaining the accuracy). However, the authors of~\cite{YangArXiv2018} focus on extracting knowledge from a trained (teacher) network and guide another (student) network in an individual training process. Furthermore,~\cite{ShenArXiv2016} used a slightly modified version of knowledge distillation for the task of pedestrian detection. In this work, we use a teacher-student architecture, but rather to leverage unlabeled data for sensorimotor control.

\section{Sensorimotor Control Across Weathers}\label{sec:method}

\begin{figure}
  \centering
  \includegraphics[width=0.8\linewidth]{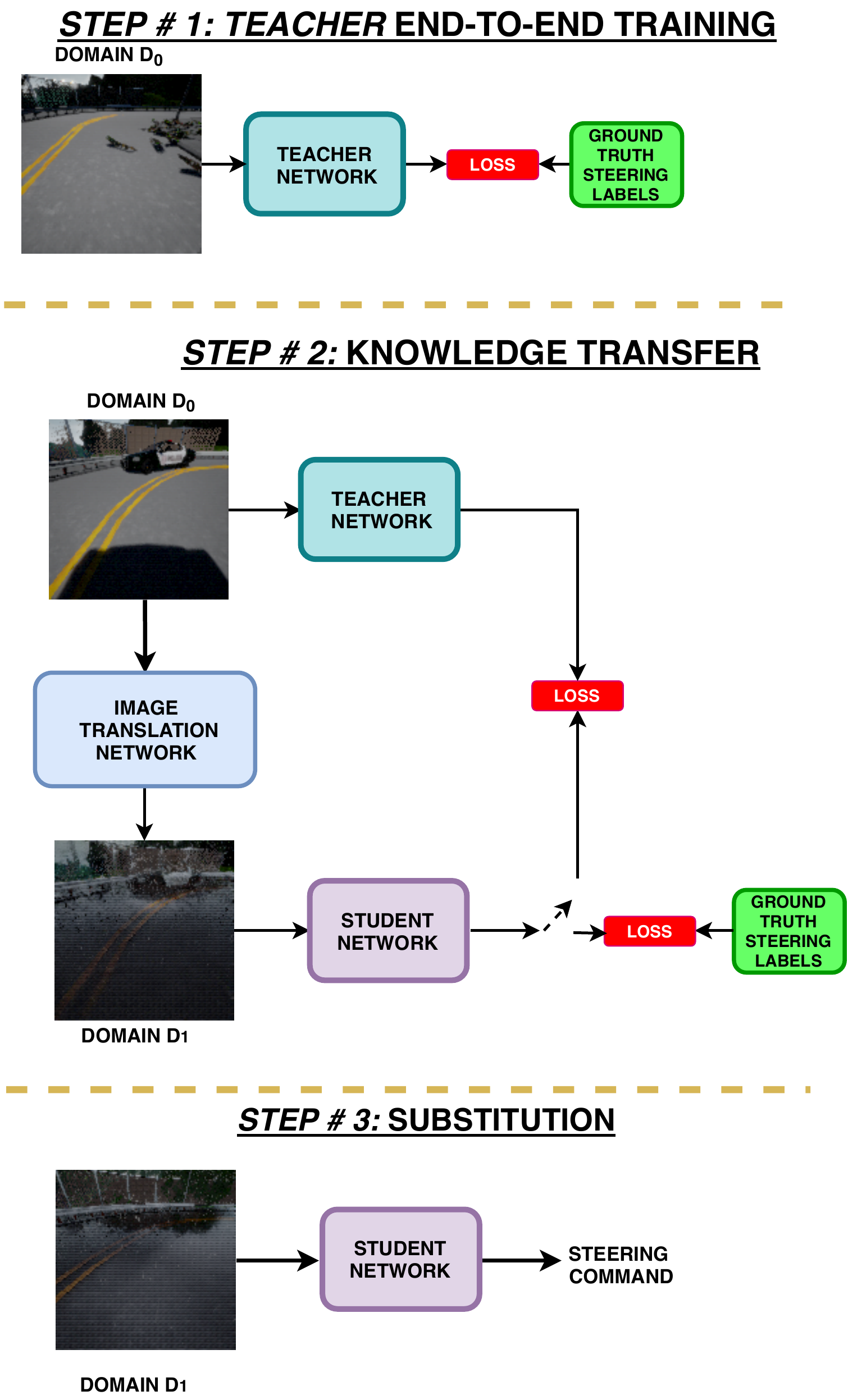}
  \caption{This figure gives a high level overview of the 3 steps for transferring knowledge between two domains $D_0$ and $D_1$ for the  purpose of sensorimotor control. Ground truth steering data for only a limited number of images from domain $D_0$ is available. Details of the framework are provided in Section \ref{sec:method}.}
  \label{fig:highlevel}
\end{figure}

In this section, we introduce a computational framework for transferring knowledge of ground truth labels from one weather condition to multiple different scenarios without any semantic labels and additional human labeling effort. Figure~\ref{fig:highlevel} gives a high-level overview of the framework.  

\subsection{Teacher End-to-End Training} 
\begin{figure}
  \centering
  \includegraphics[width=\linewidth]{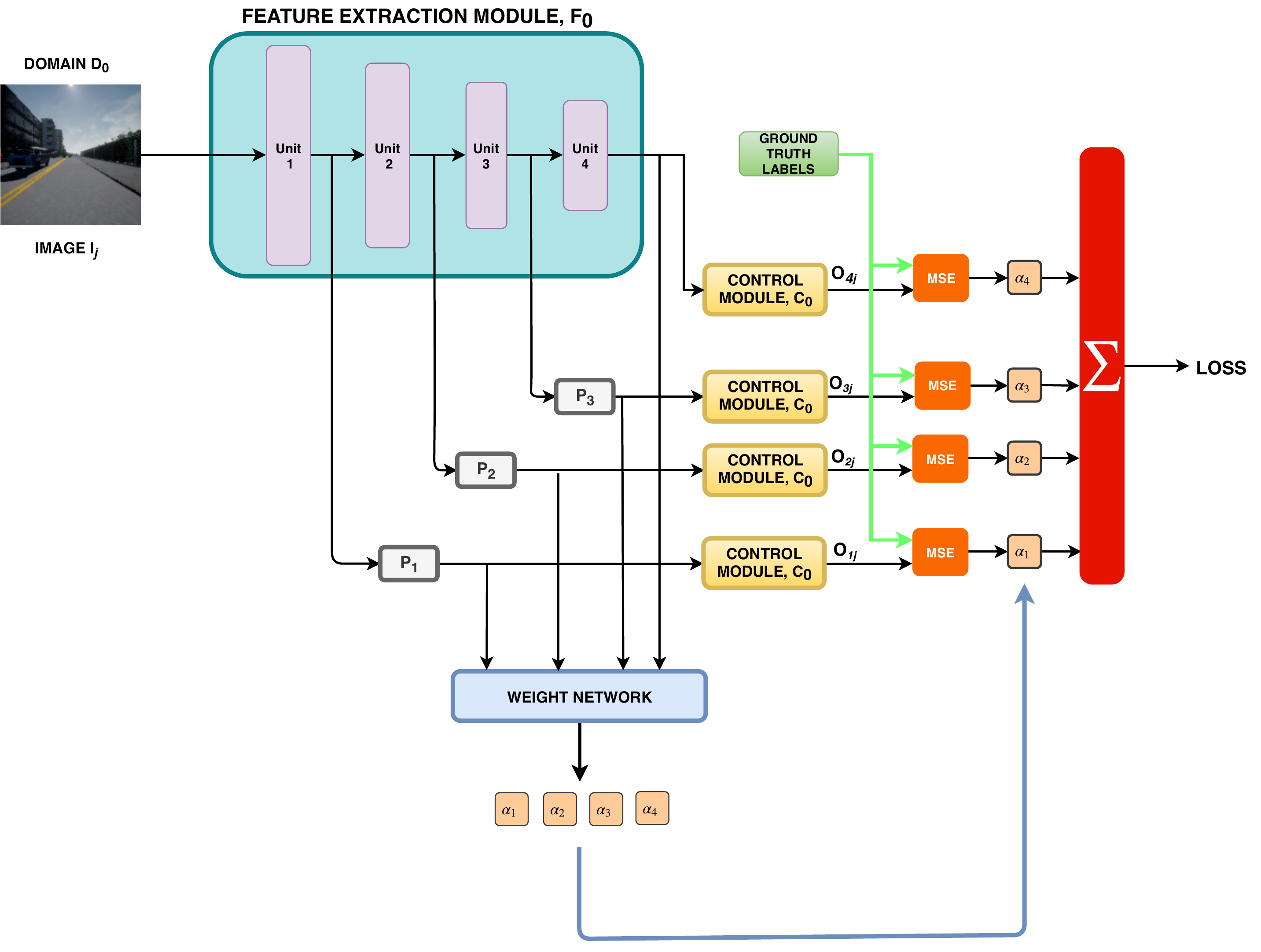}
  \caption{The figure depicts the general architecture of the model comprised of the FEM and the auxiliary control modules.}
    \label{fig:end2end}
\end{figure}

In this step, the teacher model is trained end-to-end in a supervised manner to predict the steering command of the vehicle from the raw RGB images generated by the camera placed at the front of the ego-vehicle. The training data is collected by an expert driver only once for that particular weather scenario. We refer to the images recorded under the weather condition under which this data was collected as belonging to domain $D_0$. Note that the teacher model is itself divided into a Feature Extraction Module (FEM), $F_0$ and a control module, $C_0$. The raw image (belonging to $D_0$) is passed through $F_0$ to retrieve a lower-dimensional feature representation. This feature representation is in turn fed to the $C_0$ which predicts the steering angle. A depiction of the model is shown in Figure~\ref{fig:end2end}. The FEM, $F_0$ is a sequential combination of 4 units where each unit comprises a convolutional, pooling, and activation layer. The output of unit 4 is flattened to a size of 800, which is in turn fed as an input to the module, $C_0$. The control module, $C_0$ is a series of fully connected layers and outputs the steering command. 

\noindent{\textbf{Auxiliary network.}} It might be the case that the amount of images with labels is limited or the model is too large for the task at hand. Hence, the model may tend to overfit. Therefore, during training, to mitigate the effect of overfitting, $F_0$ additionally uses auxiliary networks connected to its intermediate layers~\cite{SzegedyCVPR2015}. Each of the auxiliary networks has a control module, $C_0$ with shared weights. The projection layers, $P_1$, $P_2$ and $P_3$ project the feature maps of the intermediate layers to the dimension of $C_0$ \emph{i.e.} 800. The overall output of the teacher model is the weighted sum of the outputs of the auxiliary networks. The loss is also described by a weighted combination of the individual losses of the 4 auxiliary networks. The loss for each of the control modules is the mean squared error (MSE) between the ground truth label provided by the expert and that predicted by $C_0$. The overall loss is a weighted sum of the losses from each of the 4 control modules.

\begin{align*}
  \mathcal{L} &=\sum_{i=1}^{4} \alpha_i \cdot \mathcal{L}_{i},
  &\text{s.t.} \sum_{i=1}^{4} \alpha_i = 1,
\end{align*}

where $\alpha_i$, and $\mathcal{L}_i$ are the weighting and the error for the auxiliary network, obtained from the intermediate unit $i$ of the FEM $F_0$. The error functions are calculated as follows:

\begin{equation*}
  \mathcal{L}_i = \frac{1}{N}\sum_{j=1}^{N}(y_j - O_{ij})^2,
\end{equation*}

where $y_j$ is the ground truth steering angle obtained from the expert driver for a sample $j$ and $N$ denotes the number of total samples. $O_{ij}$ is the output of the control module corresponding to the $i$th auxiliary network for the $j$th sample. 

The weights $\alpha_i$ are themselves learned by a separate weight network. The auxiliary network that has the greatest contribution towards the overall output would also have the highest relative weight. This is important in case of limited data, wherein not all layers may be essential to train the model. In such a case the weights of the shallower auxiliary networks would be higher in comparison to the deeper auxiliary networks. Hence, a significant contribution towards the overall prediction would come from these shallow layers, thereby making the deeper layers effectively dormant. An extreme case would be when the labeled data is so small that even the first layer is enough to give a correct model prediction. In such a case, only $\alpha_1 = 1$ and all other $\alpha_i = 0, \ \text{for}\ i = 2,3,4$.

\subsection{Knowledge Transfer} 

As described in step 2 of Figure~\ref{fig:highlevel}, knowledge of ground truth labels from domain $D_0$ is transferred to domain $D_1$ using a teacher-student architecture. The output of the auxiliary networks acts as the teacher to provide supervised information to train the student.

We use the FEM, $F_0$ and control module, $C_0$ (combined, referred to as teacher) trained on images belonging to domain $D_0$, for which ground-truth steering labels are available, to transfer knowledge to a different combination of FEM, $F_1$ and control module, $C_1$ (referred to as student) for domain $D_1$, for which we  have access to only unlabeled images. The subsequent procedure is detailed in the following steps:

\begin{enumerate}
\item Image $I_0$ belonging to domain $D_0$ is passed through an image-translation-network to generate image $I_1$ belonging to domain $D_1$ in a manner that only the semantic information is preserved but the weather condition is modified. \cite{ZhuICCV2017, HuangECCV2018, MinjunECCV2018} describe methods for training a translation network in an unsupervised manner using generative adversarial networks (GANs). We use \cite{ZhuICCV2017} for our experiments. A positive implication of using these networks is that they preserve the semantics of the scene and hence the steering angle label would also be the same.
\item \noindent{\bf Hard loss:} If $I_0$ happens to have a ground truth (\emph{hard}) label then the weights of the student network are updated with these labels and the loss is referred to as the \emph{hard loss}. \noindent{\bf Soft loss:} Otherwise, a forward pass can also be done by passing $I_0$ through the teacher. Meanwhile, the corresponding image $I_1$ is passed through the student network. The output of the teacher can then used as a soft target for updating the weights of the student via the soft loss. The overall loss is the weighted average of the soft and hard losses. The weights indicate the relative importance given to the soft targets in relation to the ground truth labels.
\end{enumerate}

Note that the student network can be fed not only images from domain $D_1$ but rather multiple domains including domain $D_0$. Hence, the student network would not only be capable of predicting the steering for multiple domains but would act as a regularizer for better generalization (See P1 in Section~\ref{sec:discussion}).

\subsection{Substitution}
This refers to step 3 described in Figure \ref{fig:highlevel}. At inference time, the teacher network can be substituted with the student network to predict the correct steering command on images from all domains which the student encountered during training.

\section{Experiments}\label{sec:experiments}

In this section, we evaluate our approach on the CARLA simulator~\cite{DosovitskiyCoRL2017} version 0.8.2. It provides a total of 15 different weather conditions (labeled from 0 to 14) for two towns, \emph{Town1} and \emph{Town2}, respectively.

\subsection{Evaluation Metrics}

Finding appropriate evaluation metrics is rather challenging for navigation and driving tasks. There is no unique way to quantify these tasks. The authors of~\cite{AndersonArXiv2018} discuss different problem statements for embodied navigation and present based on these discussions evaluation metrics for some standard scenarios. In~\cite{CodevillaECCV2018}, a more extensive study on evaluation metrics for vision-based driving models is carried out. In particular, they analyzed the difference between online and offline evaluation metrics for driving tasks. The preliminary results showed that driving models can have similar mean squared error (MSE) but drastically different driving performance. As a result of this, it is not straight forward to trivially link offline to online performance due to a low correlation between them. Nevertheless, the authors of \cite{CodevillaECCV2018} found that among offline metrics not requiring additional parameters, the mean absolute error between the driving commands and that predicted ones yields the highest correlation with online driving performance. 

In addition to using this offline metric, we evaluate the online performance of the models when executing multiple and diverse turnings around corners, since it is a much more challenging task in comparison with simply moving in a straight line. The online performance is tested on the CARLA simulator across all the 15 weather conditions. For each weather condition, we evaluate the models for multiple different turns. In all experiments, the starting positions of the vehicle is just before the curve. The duration of the turn is fixed to 120 frames because it covers the entire curvature of the turn. We report the percentage of time the car remains within the driving lane as a measure of success. 

\subsection{Dataset}
For collecting ground truth training data, we navigate through the city using the autopilot mode. To demonstrate the superiority of our method, we collect a limited sample size of 6500 images for weather condition 0 of which only 3200 are labeled with ground truth steering commands. Using our proposed method we aim to transfer knowledge to the remaining 14 weather scenarios. Also, note that none of the 6500 images have any semantic labels.

The 3200 sample images with ground truth data are only available for \emph{Town2}, whereas all the offline and online evaluations are performed on \emph{Town1}. To focus the attention on the effectiveness of our approach and preserve compatibility with prior work~\cite{BojarskiArXiv2016, XuCVPR2017, CodevillaECCV2018}, the models are trained to predict the steering angle of the car while keeping the throttle fixed. The steering angles in CARLA are normalized to values between -1 and 1. The corresponding degrees for these normalized values depends on the vehicle being used. The default vehicle which we use for our experiments has a maximum steering angle of \SI{70}{\degree}. 

\subsection{Models}\label{subsec:models}
\begin{figure}
  \centering
  \includegraphics[width=0.9\linewidth]{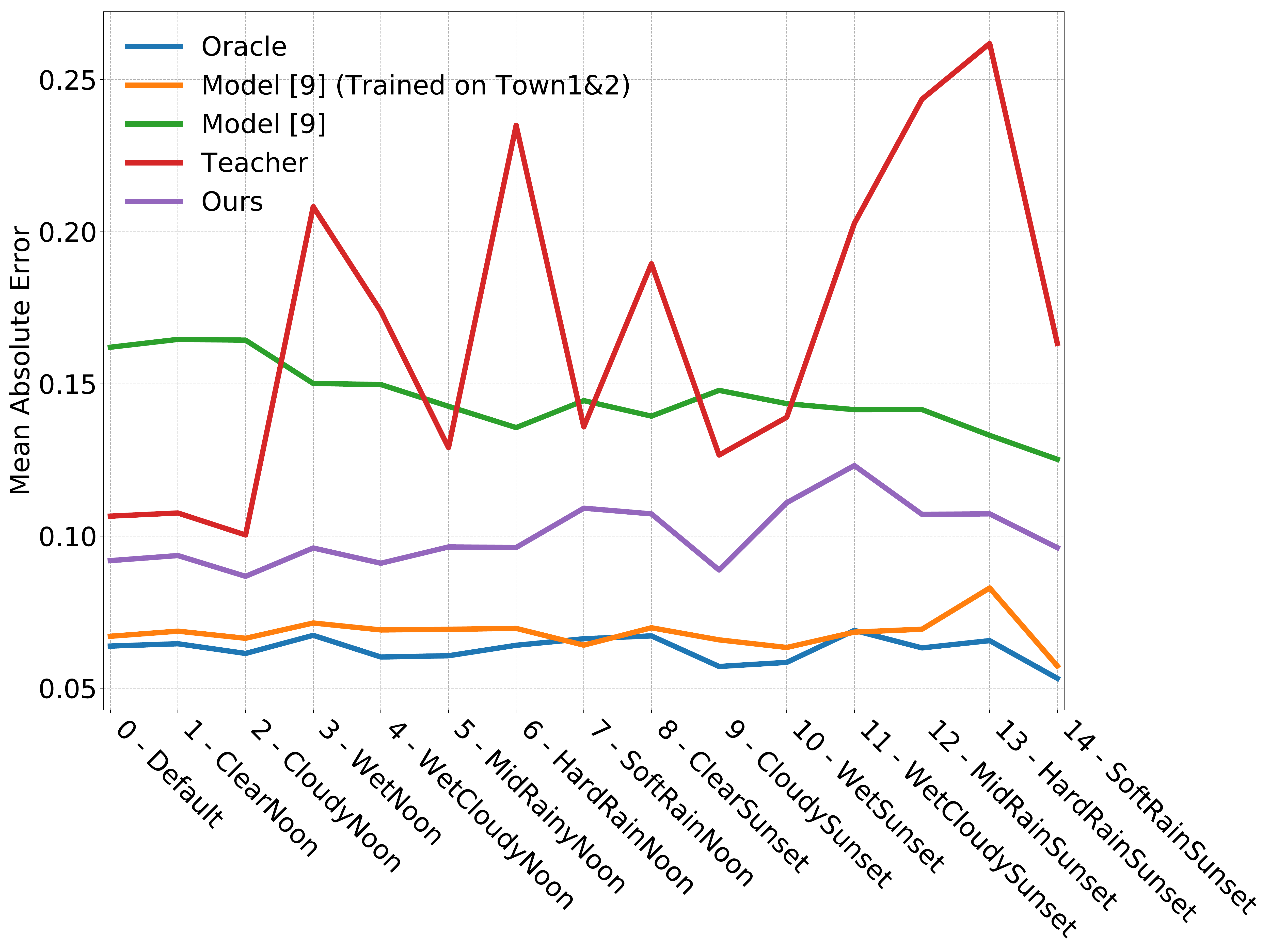}
  \caption{This plot shows the mean absolute error between the actual steering angle and that predicted by the 5 different models (see subsection \ref{subsec:models}) on data collected across the 15 different weather conditions on \emph{Town1}. Lower is better.}
  \label{fig:l1errorplot}
\end{figure}

The offline and online performance of the models described in this section are given in Figure~\ref{fig:l1errorplot} and Table~\ref{tab:turns}, respectively. Figure~\ref{fig:l1errorplot} shows the plot of the mean absolute error between the actual steering command and that predicted by all of the models. Table~\ref{tab:turns} contains the percentage for which the ego-vehicle remains within the driving lane while making turning maneuvers executed by the models across the 15 weather scenarios.

\noindent{\bf Oracle: Steering labels for all weathers. }Here we have assumed that we have access to the ground truth steering commands across all the 15 different weather conditions for \emph{Town1}. Since we are also evaluating the models on \emph{Town1} across all the weather conditions, we find in both the offline and online evaluation metrics that this model achieves the highest accuracy and hence it could serve as an upper bound for evaluating the other models along with our approach.

\noindent{\bf Model~\cite{WenzelCoRL2018}: Steering and semantic labels for weather 0. }Here we adopt the approach of~\cite{WenzelCoRL2018}, wherein the semantic labels of the images are additionally available for the 3200 labeled samples on weather 0. This additional information is used to first train what we refer to as the feature extraction module (FEM) in a supervised manner. The FEM module, in this case, is trained as an encoder-decoder architecture. The encoder encodes the input image into a lower-dimensional latent vector, while the decoder reconstructs the semantic map of the image from the latent vector. The latent vector is then used to train the control module from the ground truth steering labels. The FEM and control modules are hence trained independently and without any auxiliary networks. This FEM trained on the semantics of weather 0 is used as a teacher to train the student which is capable of producing the semantics of all the other 14 weather conditions. The authors of~\cite{WenzelCoRL2018} used the method of~\cite{ZhuICCV2017} and provide 10 separate networks for translating from weather 0 to weathers 2, 3, 4, 6, 8, 9, 10, 11, 12, and 13, respectively. The translated images for each of the 10 weather conditions along with weather 0 are fed in equal proportion to train the student. We would particularly like to evaluate our method which does not have access to any semantic labels against this model. In addition to this, we also evaluate the performance of this method on the model provided by the paper, which was trained with more than 30000 samples from both \emph{Town1} and \emph{Town2}. The performance of this model on \emph{Town1} is far superior since it was trained on much greater data and also had access to ground truth data from \emph{Town1}.

\noindent{\bf Teacher: Steering angles for weather 0. }This model is trained using only the available labeled data for weather 0 in an end-to-end manner. This model has a poor performance for the unseen weather conditions, particularly for conditions 3-14, which are considerably different in visual appearance compared to weather 0. Nevertheless, despite the poor performance this model can be used as a teacher to train the student for predicting the correct steering angles for weather conditions 1-14 for which no ground truth data exists. This approach is described in the next model. Also, note that the unlabeled data remains unutilized here.

\noindent{\bf Ours: Steering angles for weather 0. }This model is trained using the method described in Section~\ref{sec:method}, wherein knowledge is transferred from the teacher network trained on images and ground truth steering commands from weather 0 to the student network which is capable of handling images from all weathers 0-14. For a fair comparison against the model trained with semantic labels (Model~\cite{WenzelCoRL2018}, described earlier) we use the same data and generative models to translate even the unlabeled images to weathers 2, 3, 4, 6, 8, 9, 10, 11, 12, and 13, respectively. These generated images can then be fed to the student model for predicting the correct steering angles for all the 15 weather conditions.

\begin{table*}
\begin{center}
\resizebox{\linewidth}{!}{
\begin{tabular}{|l||l|l|l|l|l|l|l|l|l|l|l|l|l|l|l|l||l|}
\hline
& & \multicolumn{16}{c|}{Weather Conditions}\\
\cline{3-18}
Method & Trained on & 0 & 1 & 2 & 3 & 4 & 5 & 6 & 7 & 8 & 9 & 10 & 11 & 12 & 13 & 14 & overall \\
\hline\hline
Oracle & \emph{Town1} &  99.79 & 99.90 & 100 & 97.40 & 98.96 & 99.27 & 98.13 & 98.85 & 98.27 & 99.90 & 99.27 & 96.35 & 93.85 & 93.96 & 96.35 & 98.02\\
Model~\cite{WenzelCoRL2018} & \emph{Town1\&2} & 99.06 & 93.44 & 98.85 & 98.75 & 97.92 & 98.23 & 97.60 & 96.56 & 91.15 & 96.04 & 97.29 & 95.00 & 94.69 & 82.08 & 95.41 & 95.47 \\
Model~\cite{WenzelCoRL2018} & \emph{Town2} & 68.33 & 67.71 & 50.00 & 71.77 & 67.40 & 64.38 & 63.85 & 63.65 & 61.88 & 71.35 & 51.35 & 67.50 & 58.33 & 61.67 & 66.98 & 63.74\\
Teacher & \emph{Town2} & 92.19 & 92.40 & 82.12 & 44.38 & 51.77 & 73.65 & 32.50 & 61.56 & 49.48 & 80.10 & 60.63 & 48.54 & 35.20 & 34.27 & 50.52 & 59.29 \\
Ours & \emph{Town2} & 93.96 & 95.21 & 81.25 & 99.90 & 100 & 94.17 & 90.42 & 79.69 & 77.19 & 86.77 & 84.58 & 65.63 & 68.54 & 58.44 & 80.73 & 83.77 \\
\hline\hline
Ours (Auxiliary network 1) & \emph{Town2} & 93.96 & 93.44 & 80.73 & 92.40 & 100 & 99.69 & 90.42 & 80.10 & 77.19 & 87.50 & 91.98 & 67.40 & 66.25 & 57.29 & 81.15 & 83.97\\
\hline
\end{tabular}
}
\end{center}
\caption{This table shows the percentage for which the ego-vehicle remains within the driving lane while executing a turn for the models across the 15 different weather scenarios on \emph{Town1}. Higher is better.}
\label{tab:turns}
\end{table*}

\section{Discussion}\label{sec:discussion}

In this section, we discuss some critical insights on the experimental observations we obtained while evaluating the models. Here are some points we found worthwhile to provide some commentary based on the results provided in Figure~\ref{fig:l1errorplot} and Table~\ref{tab:turns}.

\noindent{\bf P1 - Better regularization:} It is interesting to observe that the teacher model, trained only on the available 3200 labeled samples from \emph{Town2} on weather 0 has a worse offline performance for \emph{Town1} on weather 0 in comparison to our method. This seems to imply that our approach which has been trained on multiple kinds of weather has better generalization capabilities and can even outperform its teacher when evaluated in a different town. Hence, an additional positive consequence of training the student with generated images from multiple diverse domains is that it acts as a regularizer tending to prevent overfitting to one specific domain.

\begin{figure}
  \centering
  \includegraphics[width=0.6\linewidth]{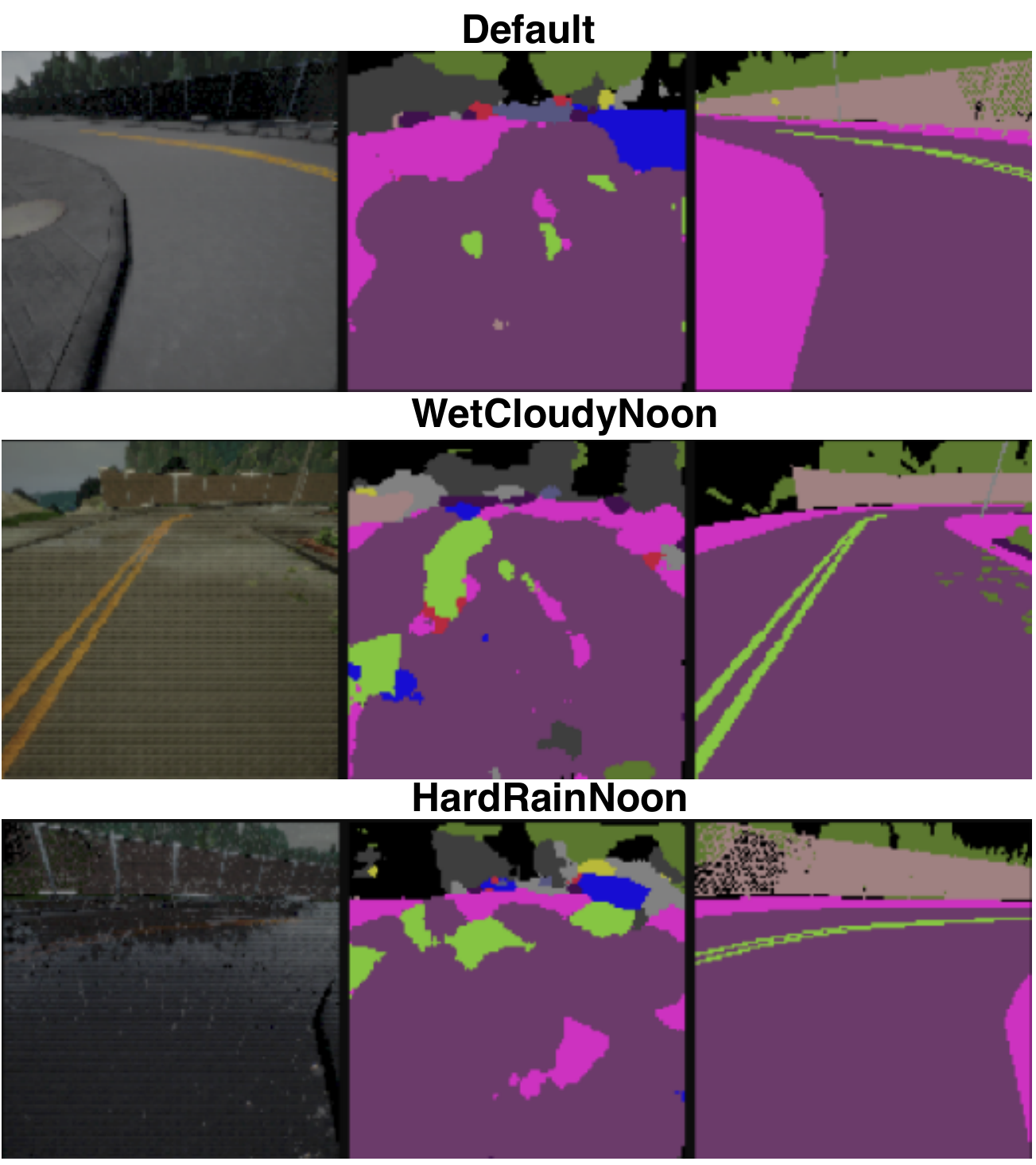}
  \caption{This plot shows three sample images (column 1) with the corresponding semantic segmentation output by the model (column 2) for 3 different weathers. The segmentation produced by the model does not reflect the actual semantic characteristics of the scene (column 3).}
  \label{fig:semantics}
\end{figure}

\noindent{\bf P2 - Semantic inconsistency:} Note that Model~\cite{WenzelCoRL2018} which in addition to having the same data and labels as our approach has also access to ground truth semantic labels. Yet, its performance is significantly poor. Upon investigation, we found that due to the limited number of semantic labels, the FEM trained as an encoder-decoder architecture seemed to be overfitting to the available data. Hence, when tested on unseen environments, the semantic segmentation output of the module breaks. The latent vector representing these broken semantics is then fed to the control module, which is incapable of predicting the correct steering command. Figure~\ref{fig:semantics} shows some sample images with the corresponding semantic segmentation outputs which are considerably different from the true semantics of the scene.

\noindent{\bf P3 - Modular training constraints:} Furthermore, the modular approach of Model~\cite{WenzelCoRL2018} wherein the FEM and control module are trained independently as opposed to an end-to-end model served to be a bottleneck in being able to learn the features universally. Also, an assumption to train the control module well is that the FEM would work perfectly well, which is not the case. Hence, the overall error of the modular pipeline would be an accumulation of the errors of the independent FEMs and control modules. We found that if we also shift the training of our approach to a modular one then performance deteriorates. This can be done in our approach by updating only the weights of the FEM of the student from the output features of the FEM of the teacher.

\noindent{\bf P4 - Auxiliary weights:} To prevent overfitting of the models, trained on limited data we used a weighted sum of the outputs of the auxiliary layers. The weights themselves were learned as part of the training. Once training of our student model was complete, we found that more than \SI{97}{\percent} of the weight was held by the first auxiliary network. This seemed to imply that only the first unit of the FEM is enough for predicting the steering command. Hence the remaining unit layers are not providing any additional information for the model. So we evaluated our model based on the output of the first auxiliary network rather than on the weighted sum of the 4 auxiliary networks. The online evaluation of this approach is given in Table~\ref{tab:turns} against the row labeled \emph{Ours (Auxiliary network 1)}. It is interesting to note that this approach is comparable in its performance with the original one. Therefore, at test time we can prune the network to a smaller size by making predictions only based on the first auxiliary network and removing the remaining 3 auxiliary networks. This would result in less computation and faster inference. 

\begin{figure}
  \centering
  \includegraphics[width=0.85\linewidth]{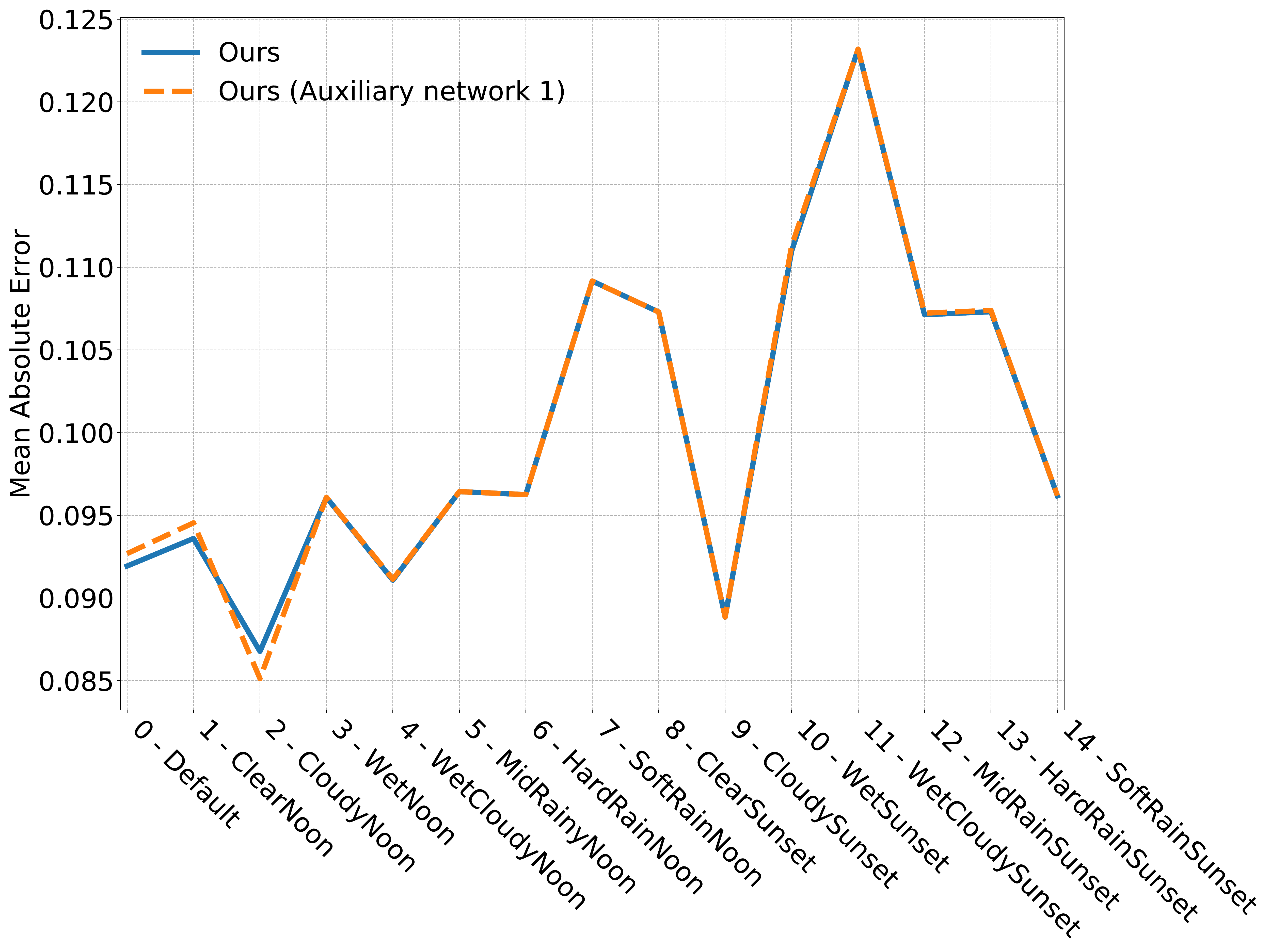}
  \caption{This plot shows the mean absolute error between the ground truth steering label and that predicted by the two models. The \textcolor{new_blue}{\textbf{blue}} curve is the weighted sum of all the 4 auxiliary networks of our model. The \textcolor{new_orange}{\textbf{orange}} line depicts the output of only the first auxiliary network of our model.}
  \label{fig:auxillaryUnit1}
\end{figure}

\noindent{\bf P5 - Online vs. offline evaluation:} Figure~\ref{fig:auxillaryUnit1} shows an offline evaluation of the two variations of our method described in the previous point across the 15 weather conditions. Note that apart from weather 0, 1, and 2, the two curves are indistinguishable from one another. However, the online evaluation results do not correspond with this observation. For weathers 3, 5, 7, and 9-14 the online performance is different despite having the same offline metric. This confirms the intuition presented in~\cite{AndersonArXiv2018} and the problems associated with evaluating embodied agents in offline scenarios. The topic of finding a correlation between offline evaluation metrics and online performance has therefore recently started to receive positive traction. It is therefore important to come up with a universal metric for evaluating various algorithms across the same benchmark. Due to the non-existence of such benchmarks, we created our own for the evaluation of the different approaches.

\begin{figure}
  \centering
  \includegraphics[width=0.5\linewidth]{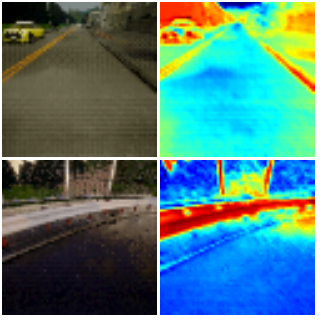}
  \caption{This figure shows the sum of the activation maps displayed as a heatmap of the first unit of the FEM of the student model for a sample taken from 2 different weather conditions. The activation maps are more prominent in regions where there are lane markings, sidewalks boundaries, other vehicles, or barriers.}
  \label{fig:stud_filter}
\end{figure}

\noindent{\bf P6 - Activation maps:} To understand the behavior of the model, which also works with only the first auxiliary network, we took the sum of the activation maps of the first unit of the FEM of the student and displayed it as a heatmap as shown in Figure~\ref{fig:stud_filter} for a sample of 2 images. We see that the activation maps are most prominent in regions where there are lane markings, sidewalks, cars, or barriers. Knowing these cues seems to be enough for the network to take an appropriate driving decision in most of the cases. Therefore, the higher-level features determined by the preliminary layers of the model are already enough to detect these objects of interest.  

\section{Conclusion}\label{sec:conclusion}
In this work, we showed how a teacher-student learning-based approach can leverage limited labeled data for transferring knowledge between multiple different domains. Our approach, specifically designed to work for sensorimotor control tasks, learns to accurately predict the steering angle under a wide range of conditions. Experimental results showed the effectiveness of the proposed method, even without having access to semantic labels as an intermediate representation between weather conditions. This framework may be extendable to other application areas for which a certain domain has ground truth data and shares a common characteristic with other domains for which no labels are available.  

\bibliographystyle{IEEEtran}
\bibliography{main.bib}  

\end{document}